Original manuscript

# Nuclear Pleomorphism in Canine Cutaneous Mast Cell Tumors – Comparison of Reproducibility and Prognostic Relevance between Estimates, Manual Morphometry and Algorithmic Morphometry

Andreas Haghofer [1,2], Eda Parlak [3], Alexander Bartel [4], Taryn A. Donovan [5], Charles-Antoine Assenmacher [6], Pompei Bolfa [7], Michael J. Dark [8], Andrea Fuchs-Baumgartinger [3], Andrea Klang [3], Kathrin Jäger [9], Robert Klopfleisch [4], Sophie Merz [10], Barbara Richter [3], F. Yvonne Schulman [11], Hannah Janout [1,2], Jonathan Ganz [12], Josef Scharinger [2], Marc Aubreville [12], Stephan M. Winkler [1], Matti Kiupel [13], Christof A. Bertram [3]

[1] University of Applied Sciences Upper Austria, Hagenberg, Austria

[2] Johannes Kepler University, Linz, Austria

[3] University of Veterinary Medicine Vienna, Vienna, Austria

[4] Freie Universität Berlin, Berlin, Germany

[5] The Schwarzman Animal Medical Center, New York, USA

[6] University of Pennsylvania, Philadelphia, PA, USA

[7] Ross University School of Veterinary Medicine, Basseterre, St. Kitts

[8] University of Florida, Gainesville, FL, USA

[9] Laboklin GmbH & Co. KG, Bad Kissing, Germany

[10] IDEXX Vet Med Labor GmbH, Kornwestheim, Germany

[11] Heska, an Antech Company, Loveland, CO, USA

[12] Technische Hochschule Ingolstadt, Germany

[13] Michigan State University, Lansing, MI, USA

Corresponding author:

Christof Bertram, Institute of Pathology, University of Veterinary Medicine Vienna, Veterinärplatz 1, 1210 Vienna, Austria. Email: Christof.bertram@vetmeduni.ac.at

# Abstract

Variation in nuclear size and shape is an important criterion of malignancy for many tumor types; however, categorical estimates by pathologists have poor reproducibility. Measurements of nuclear characteristics (morphometry) can improve reproducibility, but current manual methods are time consuming. The aim of this study was to explore the limitations of estimates and develop alternative morphometric solutions for canine cutaneous mast cell tumors (ccMCT). We assessed the following nuclear evaluation methods for measurement accuracy, reproducibility, and prognostic utility: 1) anisokaryosis (karyomegaly) estimates by 11 pathologists; 2) gold standard manual morphometry of at least 100 nuclei; 3) practicable manual morphometry with stratified sampling of 12 nuclei by 9 pathologists; and 4) automated morphometry using a deep learning-based segmentation algorithm. The study dataset comprised 96 ccMCT with available outcome information. Inter-rater reproducibility of karyomegaly estimates was low (k

= 0.226), while it was good (ICC = 0.654) for practicable morphometry of the standard deviation (SD) of nuclear size. As compared to gold standard manual morphometry (≥100 nuclei; AUC = 0.839, 95%CI: 0.701 – 0.977), the prognostic value (tumor-specific survival) of SDs of nuclear area for practicable manual morphometry (12 nuclei) and automated morphometry were high with an area under the ROC curve (AUC) of 0.868 (95%CI: 0.737 – 0.991) and 0.943 (95%CI: 0.889 – 0.996), respectively. This study supports the use of manual morphometry with stratified sampling of 12 nuclei and algorithmic morphometry to overcome the poor reproducibility of estimates. Further studies are needed to validate our findings, determine inter-algorithmic reproducibility and algorithmic robustness, and explore tumor heterogeneity of nuclear features in entire tumor sections.

Variation in nuclear size and shape of neoplastic cells is an important histologic criterion of malignancy, and various evaluation methods have been used in previous studies. The most practical method is categorical estimation by pathologists, most commonly evaluating anisokaryosis or nuclear pleomorphism. Anisokaryosis is defined as the variation of nuclear size, and nuclear pleomorphism describes the variation in nuclear size and shape. While these estimates have been shown to be relevant histologic prognostic tests for several tumors,[18,35,39,43,45] some studies suggest low inter- and intra-rater reproducibility.[46,47] For both parameters, categories (mostly three-tier such as mild, moderate, and severe) can only be vaguely defined and applications of the same thresholds between pathologists may be problematic. Another limitation is that estimates are based on categories that need to be defined arbitrarily before conducting a study, and thresholds are not based on a statistical association with patient outcome.

Alternatives to estimates are computerized measurements of nuclear size and/or shape (nuclear morphometry) in digital images, which can be either done by pathologists using a measurement software (manual morphometry) [15,19,47] or by image analysis algorithms (fully automated / algorithmic morphometry).[1,16] Morphometry can be based on two-dimensional measurements (nuclear area and shape),[32,50] or three-dimensional volume estimates based on stereological assumptions from two-dimensional histologic sections using point-sampled intercepts (volume-weighted mean nuclear volume).[15,47] Besides an assumed higher degree of reproducibility as compared to categorical estimates, morphometry enables extraction of quantitative features from histological images and thereby creates more granularity and richness of the obtained data. A potential benefit of morphometry is the output of numerical values, which allows statistical determination of meaningful prognostic thresholds at the desired sensitivity and specificity trade off. However, manual measurements by pathologists are time consuming and, thus, difficult to conduct in a routine diagnostic setting. The number of nuclei measured is 100 in most previous studies on tumors.[5,15,32,36,47,48,50] A time investment of 10-15 minutes has been reported for 75 and 166 (± 66 SD) point-sampled intercept measurements,[15,47] making this manual morphometry impracticable for routine diagnostic settings. In contrast, algorithms using state-of-the-art deep learning models are capable of eliminating human labor for these tasks and are very promising for quantitative evaluation of tumor markers.[11,38] Fully-automated nuclear morphometry can be done by post-processing of algorithmic nuclear segmentation (demarcation of all pixels representing the nuclei) masks (algorithmic output).[16]

Aside from morphometry based on nuclear segmentation, additional image analysis approaches have been used for automated evaluation of nuclear features: 1) image classification that categorizes images into tiers of anisokaryosis;[34] and 2) regression analysis that generates a continuous score based on the anisokaryosis tiers.[37] While all approaches achieve the goal of improving rater reproducibility by removing rater subjectivity, automated nuclear morphometry has several advantages as it is a quantitative and a very adaptable method. With two-dimensional morphometry, several nuclear characteristics can be evaluated individually or in combination at any desired prognostic threshold, while classification and regression approaches are restricted to the predefined classes of morphological patterns. Additionally, segmentation masks can be easily displayed as an overlay on the histological image, which allows visual verification of algorithmic performance to ensure reliability of the prognostic interpretation. Segmentation models can probably also be developed in a more consistent manner due to the nature of the ground truth data (accurate nuclear contour annotations) used for training. However, these datasets for segmentation models are more time consuming to create when compared to the other approaches that require only one label (anisokaryosis class) per image.

Canine cutaneous mast cell tumors (ccMCT) are one of the most pertinent skin tumors in dogs for possible application of these solutions due to their high frequency and malignant potential.[26] While studies on prognostic parameters for this tumor type are extensive in the veterinary oncologic literature, including the mitotic count (MC) and two multi-parameter grading systems published in 1984 and 2011,[6,9,10,12,25,27,41,44] further relatively inexpensive and practicable quantitative tests are needed to improve the prognostic ability of routine histopathologic assessment. Interestingly, despite their inclusion in the multi-parameter grading systems (e.g. karyomegaly and bizarre nuclei),[27,41] the prognostic value of histologic estimates of nuclear characteristics has rarely been investigated in ccMCT.[14,49] In an attempt to provide more objective criteria, the 2011 two-tier grading system defined a tumor as having karyomegaly if "the nuclear diameters of at least 10% of neoplastic mast cells vary by at least two-fold".[27] However, that definition of karyomegaly (origin of the word from ancient greek for "large nuclei") actually reflects anisokaryosis (variation in nuclear size), doesn't necessarily require abnormally enlarged nuclei, and would not be fulfilled when all neoplastic cells are karyomegalic. Further, a significant association of karyomegaly as a solitary parameter with survival has not been shown to date.[14] The prognostic value of manual nuclear morphometry of ccMCT has rarely been investigated for histologic [15] and cytologic ccMCT specimens [50] using either the two- or three-dimensional approaches. Fully automated solutions for nuclear morphometry have not been studied for ccMCT or any other tumors in domestic animals thus far, as opposed to tumors in humans.[1,16,53] In addition, the variation of nuclear characteristics between different tumor areas in histological sections (tumor heterogeneity) has not been evaluated for ccMCT. This information on tumor heterogeneity is relevant for deciding optimal sampling strategies of regions of interest used for image analysis.

The primary objectives of this study are as following:

- Explore the limitations of anisokaryosis estimates by pathologists, particularly regarding rater reproducibility.
- Develop nuclear morphometry methods that are feasible for a routine diagnostic setting, including practicable manual morphometry and automated morphometry.
- Investigate the measurement accuracy, reproducibility, and prognostic utility of the developed nuclear morphometry methods in comparison to the current gold standard morphometry method, anisokaryosis estimates and the MC as an independent benchmark.

A secondary objective was to evaluate the heterogeneity of nuclear size in different tumor areas.

## Material and Methods

### Study datasets

Two separate sets of histological images of ccMCT with associated data were used in this study: 1) a dataset with survival outcome and 2) a ground truth dataset with ground truth annotations for outlines of tumor nuclei,. The outcome dataset was used to determine the reproducibility and prognostic value of the nuclear evaluation methods (see below). The ground truth dataset was primarily used to train, validate and test the deep learning-based algorithm (fully automated morphometry) and was additionally used to evaluate the measurement accuracy of the nuclear evaluation methods (see below). For both independent datasets, one representative tissue block for each ccMCT (all tumors had confirmed dermal involvement with possible subcutaneous infiltration) was selected and histological slides were routinely produced using a section thickness of 2-3 µm and staining with hematoxylin and eosin. Digitization of the glass slides was done with the Pannoramic Scan II (3DHistech, Hungary) whole slide image scanner at default settings with a scan magnification of 400x (resolution of 0.25 µm / pixel). Each whole slide image (WSI) represented a different ccMCT case. Using the software SlideRunner,[2] a variable number of ROI within each WSI (1-2 for the ground truth dataset, 3-5 for the outcome dataset, see below) were cropped and exported as TIFF files using lossless compression. Each ROI had a size of 0.1185 mm² (equivalent to 0.5 standard high-power fields [38]) and an aspect ratio of 4:3. The ROI selection at low magnification was designed to include an arbitrary tumor region without paying particular attention to nuclear characteristics. This means that ROI selection was semi-random, except that tumor regions with widespread necrosis, severe inflammation and poor cell preservation and non-tumor regions were excluded, which was confirmed at high magnification once the region was selected. We decided against selecting a hotspot tumor location in this study to avoid any bias of oversampling regions with a particular nuclear characteristic (i.e., over-representing a specific morphometric parameter). Selection of hotspot locations would have also hindered analysis of tumor heterogeneity.

The small ROI size was selected to ensure a quick (diagnostically practicable) computational time for algorithmic morphometry and to allow complete ground truth annotations (all neoplastic nuclei per image) of many cases. The variable number of selected ROI for the two datasets (1-2 vs. 3-5) is justified by the distinct requirements for the use of the datasets within this study. The total number and size of ROIs used for the ground truth dataset were limited by the time investment required for the manual annotations. Considering this requirement for the ground truth dataset, we tried to obtain a realistic variability in the ccMCT images for the ground truth dataset by including a large number of different cases (and thus a low number of relatively small ROIs per case). For the outcome dataset we wanted to obtain insight into the variability of nuclear parameters between different tumor regions (tumor heterogeneity) and its effect on prognostication for the outcome dataset, and thus a higher number of ROIs from different tumor locations were selected. The higher number of ROI in the outcome dataset was also used to account for low cellularity in a few cases and ensure that at least a few hundred nuclei were available for morphometry (see below).

**Outcome Dataset**

The outcome dataset consisted of 96 cases (one tumor per patient) with known follow-up on patient survival. Histological sections were processed at Michigan State University (MSU) following routine protocols. Information on patient follow-up (date of surgery, date of death and suspected cause of death based on the clinical interpretation of the patient) was collected through a survey sent to the submitters of the surgical tissue samples. Cases were excluded from the outcome population if the patients were lost to follow-up before 12 months after surgical excision of the ccMCT or were treated by systemic or radiation therapy, i.e., all included patients were exclusively treated by surgical removal of the tumor. Following the routine trimming protocol at MSU and tumor margin evaluation (usually evaluating at least 4 peripheral margins and the deep margin), complete surgical excision was confirmed for all included cases. Postmortem examinations to verify the tumor burden were not available for any case.

For larger tumors (N = 91), 5 ROI in different tumor locations at the periphery and center were selected. For smaller tumors, in which the 5 ROI could not be placed without overlap (N = 5), the maximum number of non-overlapping ROI (3, N = 1; or 4, N = 4) was chosen.

**Ground Truth Dataset**

CcMCT cases were retrieved from the diagnostic archives of four veterinary pathology laboratories (MSU, N = 21; The Schwarzman Animal Medical Center New York, N = 14; Freie University Berlin, N = 14; Vetmeduni Vienna, N = 15) with equal numbers of high- and low-grade tumors according to the 2011 histological grading system.[27] For the samples from MSU, two ROI per WSI were used, and for the samples from other laboratories, one ROI per WSI was used. As the cases from MSU were the target domain for application of the algorithm (see outcome dataset), we decided to use one ROI from a central region and one ROI from a peripheral tumor region in an attempt to capture potential intra-tumoral variability. The other laboratories (with 1 ROI per case) were used to increase robustness of the derived algorithm regarding laboratory-derived domain shift.

Using the software SlideRunner,[2,22] one author (EP) carefully delineated the contours of the nuclear membrane of all mast cells present in these 85 images using the polygon annotation tool. The final dataset comprised 40,542 ground truth annotations (median per ROI: 455, range: 107 – 1049). The images were randomly assigned to the algorithm training subset (N = 61), validation subset (N = 11) and the test subset (N = 13), while the two images per case from MSU were always assigned to the same training or validation subset. The test subset comprised 6,111 annotations. This dataset (images and ground truth annotations) was made publicly available for research purposes on https://git.fh-ooe.at/fe-extern/mastcell-data.

The annotations of the test subset were morphometrically measured (subsequently referred to as ground truth measurements) as listed in Table 1 and described for the gold standard manual morphometry below.

**Nuclear Evaluation Methods**

For this study, the different methods of 2-dimensional nuclear size and shape evaluation investigated encompassed:

- Current routine method: anisokaryosis estimates by pathologists using two definitions:
    - two-tier classification scheme (referred to as karyomegaly)
    - three-tier classification scheme (referred to as three-tier anisokaryosis)
- Current gold standard (benchmark) method: manual morphometry by a pathologist of at least 100 neoplastic nuclei (complete sampling of grids)
- Practicable alternative: manual morphometry of 12 representative neoplastic nuclei (stratified sampling)

- Automated solution: morphometry of all neoplastic nuclei segmented by a deep learning-based algorithm

The methods of each test are specified in the following sections. Pathologists from 9 different laboratories conducted the prognostic tests and were blinded to the assessment by the other pathologists, to the results of the other prognostic tests, and to outcome information.

**Anisokaryosis Estimates**

Eleven veterinary pathologists (TAD, CAA, PB, MJD, AFB, AK, KJ, RK, SM, BR and FYS) participated in this study and were anonymized by a random identification number (P1-11). Anisokaryosis estimates were conducted at two time points with a wash-out time of at least 2 weeks. For time point 1 (11 participants), pathologists estimated the degree of anisokaryosis using two systems (see below) for the cases of the outcome dataset. For time point 2 (9 participants), pathologists were instructed to estimate the degree of anisokaryosis a second time and to measure 12 neoplastic mast cell nuclei (manual morphometry, see below) for each tumor in the test subset of the ground truth dataset and the outcome dataset. Images were provided to the pathologists through the online annotation platform EXACT.[33] The images from the 3-5 ROI of each case of the outcome dataset were stitched to an image panel and separated by a black line to allow simultaneous viewing.

Two classification schemes (two- and three-tier) for anisokaryosis estimation were applied by the pathologists for each case at two time points, resulting in 3,840 data points for the outcome dataset and 234 data points for the test subset of the ground truth dataset. First, karyomegaly was assessed according to the definition given by Kiupel et al. for the two-tier MCT grading system,[27] "nuclear diameters of at least 10% of neoplastic mast cells vary by at least two-fold," as: 1) absent or 2) present. Secondly, the three-tier anisokaryosis system consisted of the following categories: 1) none to mild, 2) moderate, and 3) severe variation in nuclear size of neoplastic mast cells. These stratifications were intentionally vaguely defined as this is common practice in current veterinary literature.[17,18,28,30,39,40,42,46] We also wanted to avoid creating arbitrary definitions of the three categories without known association with patient outcome.

**Gold Standard Manual Morphometry of ≥100 Nuclei**

Consistent with previous literature [5,32,36,48,50] on nuclear morphometry in veterinary pathology, the gold standard morphometry method included manual annotations of at least 100 neoplastic mast cell nuclei. While acknowledging that this test is too time consuming for routine diagnostic application, we considered this method the benchmark for comparison with the other nuclear evaluation methods.

One pathologist (CAB) evaluated the images of the test subset of the ground truth dataset and ROI 1 of the outcome dataset by this method. In these images, a 5x6 grid overlay with thin black lines was added, i.e., the images were separated in 30 equally sized grid fields. Using the software SlideRunner,[2,22] the pathologist annotated as many grid fields (complete sampling) as needed to reach 100 nuclei. The sequence of grid fields selected followed a uniform meandering pattern that included the central grid fields first and the fields at the image borders last. Each grid field was completely annotated including each neoplastic nucleus that touched the grid borders and excluded nuclei that were cut off at the image borders. For the 109 evaluated images, an average of 108 annotations per image (minimum: 100; maximum: 133; total: 11,766) were created with an average of 7.7 / 30 grid fields (minimum: 2; maximum: 29). Time investment for the ≥100 annotations per image was 17.5 minutes on average (minimum: 14.5; maximum: 24).

From these manual annotations, different characteristics of the probability density function (Table 1) were calculated (morphometry). Nuclear area was defined by the number of pixels within the segmented nuclei and subsequent conversion into µm² based on the scan resolution. The SD of the nuclear area reflects the variation of the nuclear size and thus was used as the primary parameter to compare with the three-tier anisokaryosis estimates by pathologists. To approximate the two-tier system definition for karyomegaly, "nuclear diameters of at least 10% of neoplastic mast cells [i.e., proportion of large nuclei] vary by at least two-fold [i.e., extent of nuclear size difference]",[27] we evaluated two morphometric parameters. The proportion of abnormally enlarged (karyomegalic) cells was calculated by the number of nuclei above a case-independent reference size divided by the number of all tumor nuclei detected. The case-independent size for large nuclei was >50.3 µm² (two times the median area of all the annotated nuclei in the training/validation subset of the ground truth dataset) or >37.8 µm² (the 90$^{th}$ percentile of the annotated nuclei). We decided to select the thresholds for the karyomegaly definition from the ground truth dataset (and not based on statistical correlation with the outcome dataset) to avoid overfitting of the parameter on the outcome dataset. The extent of nuclear size variation was determined by the 90$^{th}$ percentile divided by the median, representing the factor by which the largest 10% of nuclei, i.e., potentially karyomegalic cells, differ from the median nuclear size of the same tumor (case-specific reference size). Instead of using the diameter according to the karyomegaly definition used in the two-tier system, we based our calculations on the nuclear area, as an increase of the diameter is not proportional to the actual increase of nuclear size and a nuclear diameter is not representative of size for oval nuclei.

As indicators of nuclear shape, we measured eccentricity and solidity, as implemented in the scikit-image framework.[52] The eccentricity measure is used to evaluate the roundness of the detected object. It is calculated by the ratio of the distance between the focal points of an ellipse over the major axis length. The closer this ratio is to 1, the more elongated the shape is. The closer the ratio is to 0, the more circular the shape is. For the calculation of the solidity measure, we used the ratio of the detected object area compared to the resulting convex hull of this area. A solidity value of 1 indicates that the detected pixel area has the same size as its convex hull. The closer this value is to 0, the more indentations are present and/or the larger the indentations are, corresponding to the bizarre nuclei definition of the 2011 grading system.[27] Nuclear indentation thresholds evaluated were <0.913, <0.936, <0.943, representing the 2$^{nd}$, 5$^{th}$, and 10$^{th}$ percentile, respectively, of the training/validation subset of the ground truth dataset. The percentage of indented nuclei over all detected nuclei was calculated. For calculation of the prognostic value of mean solidity, the direction of the values was inverted (i.e., higher degree of nuclear indentations are represented by larger values) by subtracting the mean solidity value from 1 for each case.

**Table 1.** List of parameters for manual and algorithmic morphometry evaluated in this study.

| Feature | Measurement | Parameters |
|---|---|---|
| Size | Area (in µm²) | Mean, median, standard deviation (SD), 90th percentile (90th P), 90th P / median, mean of the largest 10% of the nuclei, percentage of large nuclei (>37.8 µm² or >50.3 µm²), skewness (asymmetry of the data distribution) |
| Size | Eccentricity | Mean, SD, skewness |
|  | Solidity | Mean, SD, Percentage of nuclei with indentation (solidity <0.913), skewness |

**Practicable Manual Morphometry of 12 Nuclei**

At time point 2 (after anisokaryosis estimates), the 9 pathologists measured 12 neoplastic mast cell nuclei per case (Fig. 1). For cases with notable nuclear size differences, participants were instructed to perform stratified sampling with selection of 4 nuclei with a small, intermediate, and large area each. The relatively low number of nuclei was needed for practicability in routine diagnostics, while a stratified sampling method (see below) aimed at a representative frequency distribution. The images from the test subset of the ground truth dataset were analyzed first (resulting in a total of 1,407 annotations by 9 pathologists; in three instances, a participant had annotated one nucleus too many), and subsequently, the images of the outcome dataset were evaluated (resulting in a total of 10,368 annotations). Pathologists were asked to use the one ROI of the outcome cases with the presumed (estimated) highest degree of anisokaryosis. The polygon annotation tool in Exact [33] was used to outline the neoplastic nuclei. Based on the annotations, the mean, SD, and maximum nuclear area were calculated for each case and pathologist. We restricted the practicable manual morphometry to size parameters as pathologists reported difficulty in delineating the shape of the nuclei.

**Automated Morphometry using a Supervised Deep Learning-Based Algorithm**

We previously developed a Unet ++-based segmentation model to create binary masks where pixels corresponding to the nuclei of the initial slide image are depicted as positive foreground (ones) in front of a negative background (zeros).[55] We trained our model using the training and validation subsets of the ground truth dataset. A more detailed description of the development method of the segmentation model is provided in the supplemental material. The developed model is available through https://git.fh-ooe.at/fe-extern/mastcell-data. As a post-processing step, we used connected component labelling provided by the Scikit-Image Framework [52] to detect the individual nuclei within the segmentation mask. Based on the identified nuclei, we applied a filtering mechanism for objects smaller than ~7 µm² to exclude segmented objects that do not represent complete and valid nuclei. Subsequent to segmentation and filtering of the individual mast cell nuclei, different morphometric parameters were calculated for each case (algorithmic morphometry) as listed in Table 1 and detailed in the section "gold standard manual morphometry" above.

The developed algorithm was used to analyze the images of the test subset of the ground truth dataset and the outcome dataset. For the outcome cases, calculations were done for each ROI separately and averaged for all ROI per case. Algorithmic segmentation resulted in 5,794 objects for the test subset of the ground truth dataset (average per image: 445; minimum: 75; maximum: 929) and 176,912 objects for the outcome dataset (average per case: 1,842; minimum: 604; maximum: 4,090). Algorithmic morphometry was calculated from the uncorrected predictions, i.e., no expert review of the segmentation masks was conducted in this study. However, for application of the algorithm in a diagnostic setting we do recommend expert review of the segmented nuclei.

## Mitotic Count (MC, Prognostic Value Benchmark) and Histologic Grade

As the prognostic value of the gold standard manual nuclear morphometry (using the two-dimensional method) is not well established for ccMCT, the MC was determined as an independent benchmark with the intention of providing guidance for the interpretation of prognostic value of the nuclear evaluation methods. The MC is probably the single most important prognostic histological parameter for ccMCT with numerical values[10,26] and is valuable in understanding how well a single prognostic test can theoretically discriminate patient survival for this specific outcome dataset. This does not mean that the evaluated prognostic parameters must have a better prognostic value than the MC to be of relevance as cellular proliferation and nuclear pleomorphism reflect distinct malignancy characteristics of the tumor. We consider the histologic grading schemes [27,41] to be less appropriate benchmark tests for this study because these systems combine nuclear characteristics (i.e., karyomegaly and bizarre nuclei) with other morphologic criteria (such as the MC); thus, they are theoretically superior to a single morphometric parameter. As the grades are categorical values, some statistical tests for evaluation of the prognostic relevance (see below) are not possible as compared to numerical tests (such as the MC and nuclear morphometry), also hindering a thorough statistical comparison of the prognostic value with nuclear morphometry.

The MC was determined by one pathologist (CAB) in the WSIs of the outcome dataset according to current guidelines.[38] The software SlideRunner [2] with a plug-in for an rectangular bounding box overlay (4:3 ratio, area of exactly 2.37 mm²) was used, as previously described.[7] This area box was placed in a mitotic hotspot location, which was selected based on the impression of mitotic activity evaluated by a pathologist in several tumor areas. Regions with widespread necrosis, severe inflammation, low tumor cell density, poor cell preservation, and extensive artifacts were excluded, if possible. All mitotic figures (according to published definitions [20]) within this hotspot area were annotated by screening these areas at high magnification twice to minimize the number of overlooked mitotic figures. The number of annotations represents the number of mitotic figures.

The two-tier grade was assigned by one pathologist (CAB) for the cases of the outcome dataset according to the published criteria.[27] The same MC as determined above was used for grading and the other parameters were determined in the WSIs. For both the MC and tumor grade, the pathologist was blinded to patient outcome and results of the nuclear evaluation methods.

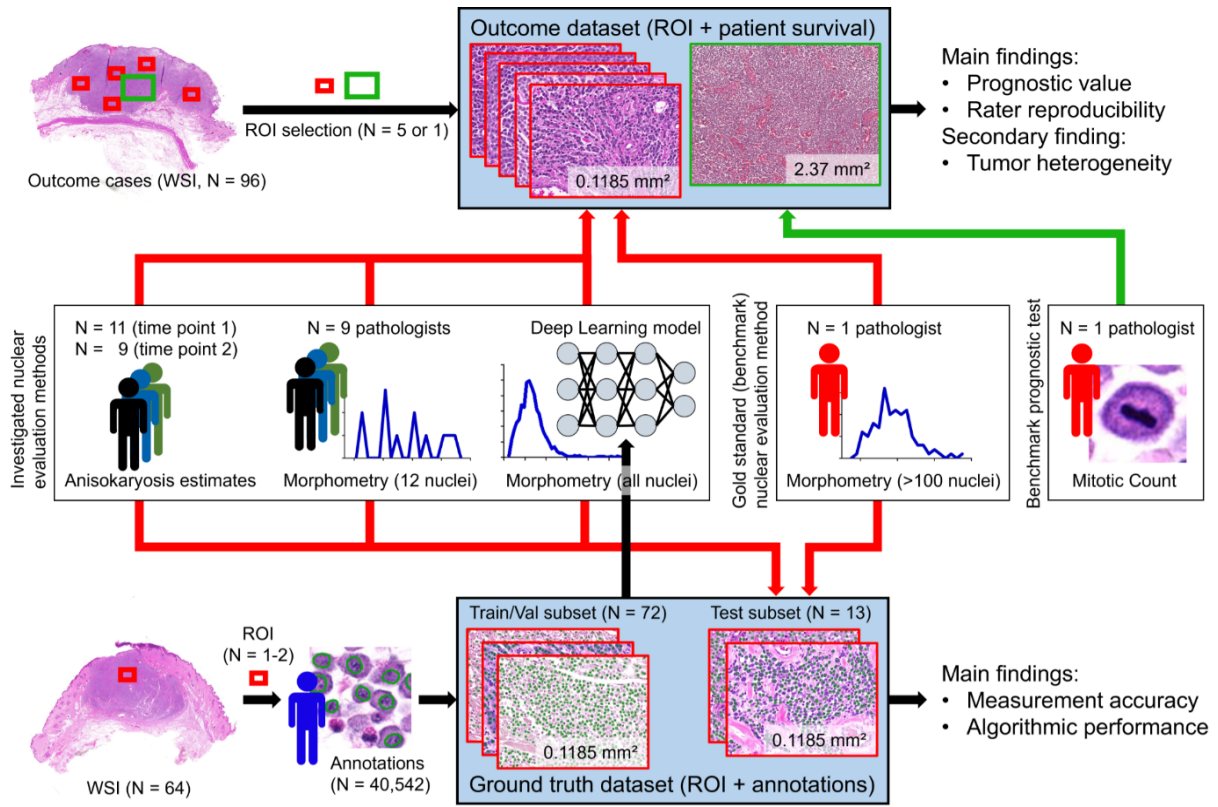

**Figure 1.** Overview of the study design. Two different sets of histological images with associated data, the ground truth dataset and the outcome dataset, were used to investigate different nuclear evaluation methods.

ROI, regions of interest; WSI, whole slide images; Train/Val subset, training and validation subset.

## Statistical Analysis

Statistical analysis and graph creation was performed by GraphPad Prism version 5.0 (GraphPad Software, San Diego, CA), IBM SPSS Statistics version 29.0 (IBM Corporation, Armonk, NY) and R version 4.2.2 (R Foundation, Vienna, Austria).

**Rater reproducibility (outcome dataset)**

Rater reproducibility was determined for the cases of the outcome dataset. For categorical estimates, inter- and intra-rater reproducibility was determined by Light's Kappa and weighted Cohens Kappa (k). The level of agreement was interpreted as poor = 0, slight = 0.01–0.20, fair = 0.21–0.40, moderate = 0.41–0.60, substantial = 0.61–0.80, and almost perfect = 0.81–1.00.[24] For pathologists' measurements (numerical values), inter-rater agreement was measured by the intraclass correlation coefficient (ICC; 2-way agreement, single measures, random) with the following interpretation: poor = 0–0.39, fair = 0.40–0.59, good = 0.6–0.74, and excellent = 0.75–1.00.[24] To compare the estimate categories assigned by each pathologist with the actual SD area (algorithmic morphometry) of the individual case, linear regression was used.

**Test accuracy (ground truth dataset) and correlation (outcome dataset)**

The test accuracy of the nuclear evaluation methods was determined on 13 images from the test subset of the ground truth dataset. Segmentation performance of the deep learning model (i.e. overlap of the algorithmic segmentation map with the area of the ground truth annotations) was determined by the Dice Score. The individual objects from the algorithmic segmentation were identified, and the algorithmic performance to detect individual nuclei as compared to the ground truth was measured by the F1 score, recall (sensitivity) and precision. Measurement errors of algorithmic and manual measurements for the entire image were determined by comparison to the ground truth measurement using the root mean squared error (RMSE). Algorithmic and manual measurements, and pathologists' estimates were compared with the ground truth measurements using scatterplots. The correlation of the algorithmic morphometric parameters was analyzed on the outcome dataset using Pearson's method.

**Prognostic value (outcome dataset)**

The outcome metrics primarily evaluated in this study (outcome dataset) were tumor-related mortality at any time of the follow up period and tumor-specific survival time. Dogs that died due to other causes (not considered ccMCT-related by the clinician) were grouped with cases that survived the follow-up period (ROC curves, scatter plots, sensitivity and specificity) or were censored (Kaplan Meier curves and hazard ratios). The bias of this outcome metric is the variable follow-up period between cases without reported death (minimum of 12 months) and the lack of conclusive proof of the cause of death. It can't be ruled out that tumor-related mortality was missed in a few cases with a relatively short follow-up period, while acknowledging that most ccMCT patients that survive the first 12 months will die from other causes. To eliminate the bias of a variable follow-up period and unproven cause of death, tumor-specific mortality and overall mortality within the first 12 months of the follow-up period were used as alternative outcome metrics. Overall death was defined as the occurrence of death regardless of the cause of death.

Numerical tests (algorithmic and manual morphometry and MC) were analyzed by receiver operating characteristic (ROC) curves (plotting sensitivity against specificity) for numerous thresholds) and the area under the ROC curve (AUC) with 95% confidence intervals (95% CI). The distribution of the algorithmic measurements and the MC were displayed in scatter plots comparing cases with tumor-related mortality and other cases. Numerical tests with an AUC ≥ 0.700 were dichotomized by thresholds, resulting in uniform sensitivity value for the different tests, which allows comparison of the associated specificity values. For the MC, the threshold proposed by Romansik et al. [44] was used to group cases with values of 0-5 and ≥6. The range of the morphometric measurements were divided into 200 intervals (thresholds increased by 0.5% steps) and two thresholds leading to a sensitivity of 76.9% (threshold 1; 10 true positive cases, TP, and 3 false negative cases, FN; represents the sensitivity value for the MC), and 53.8% (threshold 2; 7 TP, 6 FN) were selected. If multiple cut-off values resulted in the desired sensitivity value, the highest value was picked.

With the categorical data (dichotomized numerical tests and pathologists' estimates), Kaplan Meier curves, hazard ratios with 95% CI (univariate cox regression), as well as sensitivity (Sen, also known as recall), specificity (Sp) and precision (Pre, also known as positive predictive value) were calculated.

The pathologists' measurements were analyzed individually, averaged, and were combined (repeated measure data), when reasonable. For combined data, bootstrapping (AUC values) or a mixed model (cox regression) were used to calculate 95% CIs.

**Intra-tumoral Heterogeneity (outcome dataset)**

Heterogeneity was evaluated for automated morphometric measurements between the different ROI per case using the outcome dataset. The difference between the 3-5 ROI was determined by the coefficient of variation (SD / mean). The influence of the number of ROI used for prognostic evaluation was determined by the AUC calculated from the mean measurements of 1, 2, 3, 4 and 5 ROIs based on their order of selection within the WSI. Tumor heterogeneity as a prognostic test was defined by the SD between the 3-5 ROI and by the proportion of ROI with a morphometric measurement above threshold 1 (defined as a hotspot) over all evaluated ROI.

# Results

### **Rater reproducibility**

Inter-rater reproducibility was slight to fair for estimates (time point 1) of karyomegaly (k = 0.226) and three-tier anisokaryosis (k = 0.187), whereas it was good for practicable measurements (12 nuclei) of mean nuclear area (ICC = 0.637, 95%CI: 0.482 – 0.750), SD of nuclear area (ICC = 0.654; 95%CI: 0.577 – 0.730) and maximum nuclear area (ICC = 0.683, 95%CI: 0.603 – 0.756). Categorization of the practicable measurements (based on threshold 1 of the mean pathologists' values, see below) resulted in k = 0.432, k = 0.471, k = 0.497 for mean, SD and maximum nuclear area, respectively. Consensus on the categories by at least 8/9 pathologists (excluding the two pathologists that did not do the measurements) occurred in 43/96 cases (44.8%) for karyomegaly estimates, in 62/96 cases (64.6%) for mean nuclear area measurements, in 65/96 cases (67.7%) for SD of nuclear area measurements, and in 69/96 cases (71.9%) for maximum nuclear area measurements.

Intra-rater reproducibility between time points 1 and 2 was moderate for estimates of karyomegaly (k = 0.51; 95%CI: 0.44 – 0.58) and three-tier anisokaryosis (k = 0.60, 95%CI: 0.55 – 0.65). Comparing inter- and intra-rater reproducibility, the higher inter-rater inconsistency can be largely explained by different interpretations by the pathologists of the vaguely predefined thresholds. The variable application of the thresholds is illustrated in Fig. 2, which plots the three-tier anisokaryosis categories to its corresponding algorithmically measured SD of the nuclear area.

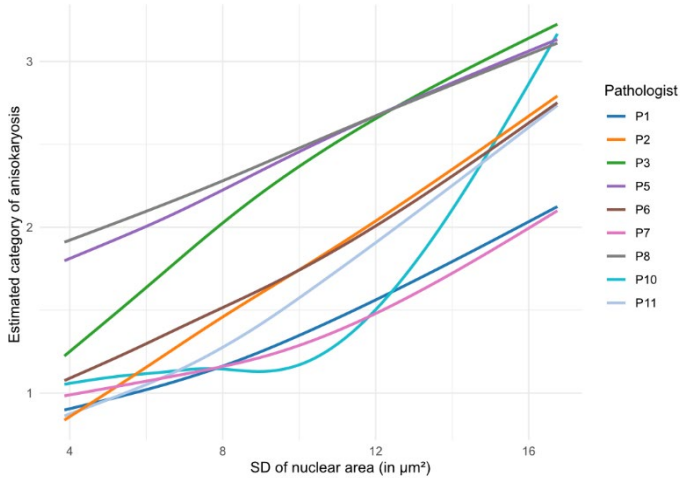

**Figure 2.** Illustration of the increase in the average three-tier anisokaryosis estimate (time point 2, outcome dataset) depending on the algorithmic standard deviation (SD) of the nuclear area for each pathologist (curves determined by linear regression). The curves show which anisokaryosis category was likely assigned by the corresponding pathologist to a case depending on the SD of nuclear area measured. Curves were smoothed using a spline regression.

### Test accuracy and correlation

Higher categories of karyomegaly and three-tier anisokaryosis estimates were more commonly assigned to cases with higher SD of nuclear area based on the ground truth annotations or based on their own measurements of 12 nuclei, however, with marked inconsistency between cases and pathologists.

Comparison of the RMSE of the gold standard manual morphometry with the other morphometry methods revealed that the practicable manual method (12 nuclei) had a markedly higher degree of error and algorithmic morphometry had a slightly higher degree of error. Overall, the measurement errors were lower for nuclear size parameters than for nuclear shape parameters.

Segmentation performance of the deep learning-based algorithm on the test subset of the ground truth dataset was Dice score = 0.785 (Fig. 3). The algorithm detected 5.794 individual neoplastic objects, which, compared to the annotations in the ground truth dataset, resulted in a performance of $F_1$ score = 0.854, recall = 0.830, and precision = 0.880. The model was able to adequately segment most tumor nuclei even in images with prominent metrachromatic cytoplasmic granules, which partially obscured the nuclei, and severe eosinophilic infiltration. Comparison of the gold standard manual measurements with the algorithmic measurements on the outcome dataset proves similar output of these methods for the nuclear size parameters, however not for the evaluated nuclear shape parameters.

Most morphometric parameters of nuclear area had a very strong correlation with each other including the 90$^{th}$ percentile to the median with a correlation coefficient of 0.936 (manual morphometry of ≥100 nuclei) or 0.970 (algorithmic morphometry). SD of eccentricity and solidity poorly to moderately correlated with the nuclear area parameters (SD, mean, median, 90$^{th}$ percentile, mean of the largest 10% and percentage of large nuclei) with coefficients ranging between 0.100 to 0.320 for manual morphometry of ≥100 nuclei and between 0.404 to 0.674 for algorithmic morphometry.

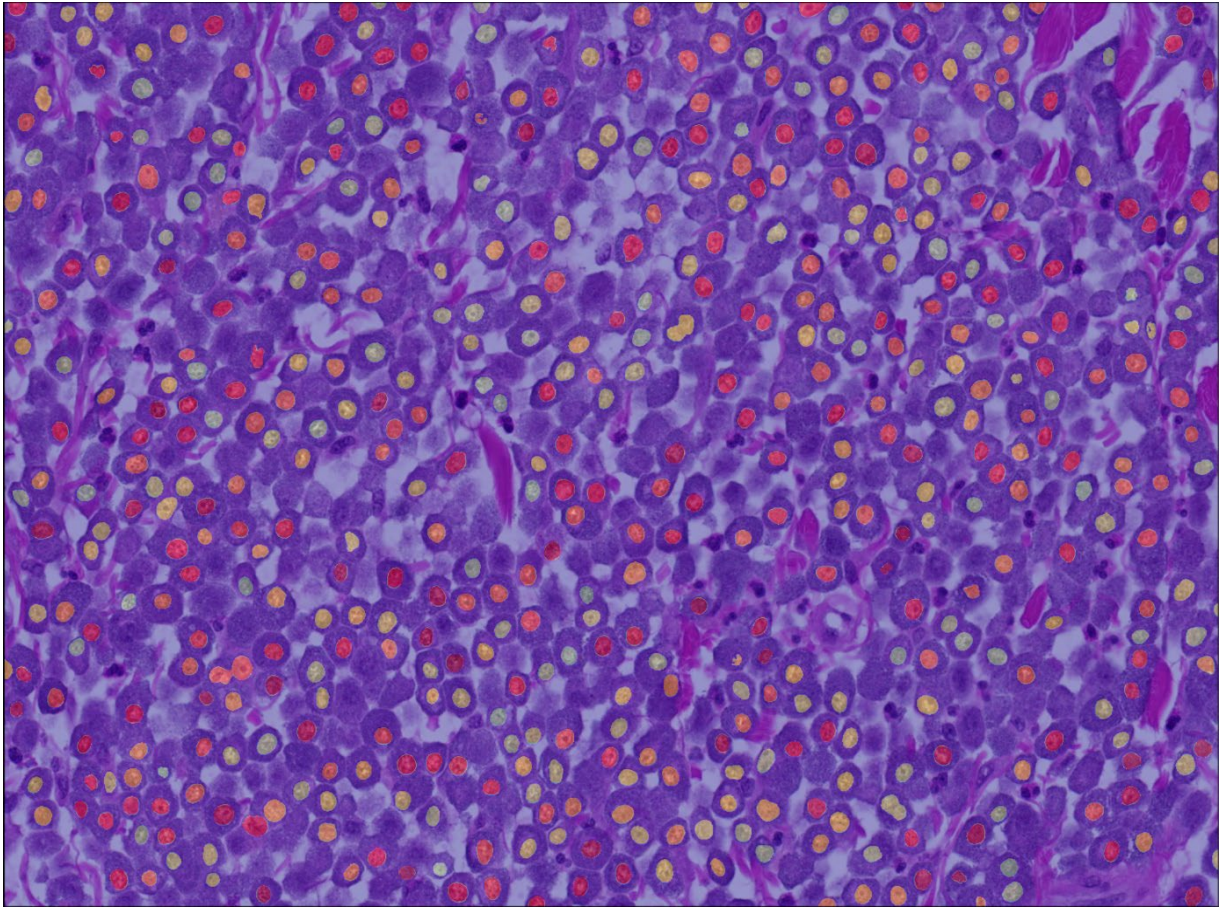

**Figure 3.** Example of an algorithmic segmentation mask for an outcome case. In rare instances two nuclei are connected (undersegmentation). The algorithm has the tendency for omission of few neoplastic nuclei, whereas rarely false objects are detected.

**Prognostic Value**

Of the 96 cases (78 low-grade cases and 18 high-grade cases) included in the outcome dataset, death was attributed to the ccMCT in 13 cases with a median survival time of 4.3 months (range: 0.5 - 24). Other cases (N = 83) were lost to follow-up (N = 72) with a median follow-up period of 24 months (range: 12 - 45.3 months) or were reported to have died due to ccMCT-unrelated causes (N = 11) with a median survival time of 8.5 months (range: 0.2 - 29.7). At 12 months after surgical removal, 10 had died due to ccMCT-related and 6 due to ccMCT-unrelated causes.

The frequency distribution of nuclear morphometry was able to discriminate cases without tumor-related and overall death from other cases as indicated by high AUC values for each method. In comparison to the gold standard, nuclear morphometry (≥100 nuclei), practicable manual morphometry (12 nuclei) and algorithmic morphometry achieved higher AUC values for patient survival (Fig. 4). For example, AUC (tumor-specific survival) of the SD of nuclear area was 0.839 (95% CI: 0.701 – 0.977) for gold standard morphometry, ranged between 0.823 to 0.960 for practicable morphometry of the individual pathologists, and was 0.943 (95% CI: 0.889 – 0.996) for algorithmic morphometry. The independent benchmark MC had an AUC for tumor-specific survival of 0.885 (95%CI: 0.765 - 1.00, $p < 0.001$), which was somewhat below most morphometric size parameters of the practicable and algorithmic approach. On the other hand, algorithmic shape measurements had AUC values below the MC, while mean shape values did not provide any prognostic information (AUC values close to 0.5).

Comparing the two outcome groups (tumor-related death vs. other) shows that the nuclear size measurements are able to distinguish patient outcome at a high sensitivity (threshold 1) or specificity (threshold 2), depending on the selected threshold (Fig. 5). The specificity and precision values based on threshold 1 are provided in Table 2 for manual and algorithmic SDs of area measurements, revealing a high performance of the algorithm and high variability between the pathologists in the threshold required to obtain the same sensitivity values. For SD of nuclear area based on algorithmic morphometry, a tumor-specific death rate of 3.9% (false omission rate) and 52.6% (precision) was determined for cases below and above the threshold of ≥9.0 µm², respectively.

The categorical anisokaryosis estimates by pathologists resulted in highly variable sensitivity (Sen) and specificity (Sp) values ranging between Sen = 100% / Sp = 4.8% (pathologist 8) to Sen = 46.2% / Sp = 98.8% (pathologist 7) for anisokaryosis 1 vs. 2 and 3, between Sen = 84.6% / Sp = 69.9% (pathologist 5) to Sen = 0% / Sp = 100% (pathologist 1) for anisokaryosis 1 and 2 vs 3, and between Sen = 92.3% / Sp = 31.3% (pathologist 11) to Sen = 0% / Sp = 98.8% (pathologist 5) for karyomegaly (Fig. 6). The performance of the estimates by almost all pathologists were below the algorithmic ROC curve (Fig. 6). As compared to the MC classified by the threshold proposed by Romansik et al. [44] (Sen = 79.6%, Sp = 92.8%), the sensitivity was slightly higher for the 2011 two-tier histologic grade (Sen = 84.6%, Sp = 91.6%) due to the combination with the nuclear characteristics.

Kaplan-Meier curves and hazard ratios determined that patient survival time was significantly different for cases with low versus high anisokaryosis based on the predefined categories of the estimates and based on threshold 1 for nuclear morphometry of SD of area (Fig. 7). The hazard ratios of the morphometric size measurements were higher than those for the shape measurements; however, both were lower than the MC (benchmark) or histologic grading system with a value of 30.5 (95%CI: 7.8 – 118.0, p < 0.001) or 46.5 (95%CI: 9.6 – 223.3, p < 0.001), respectively.

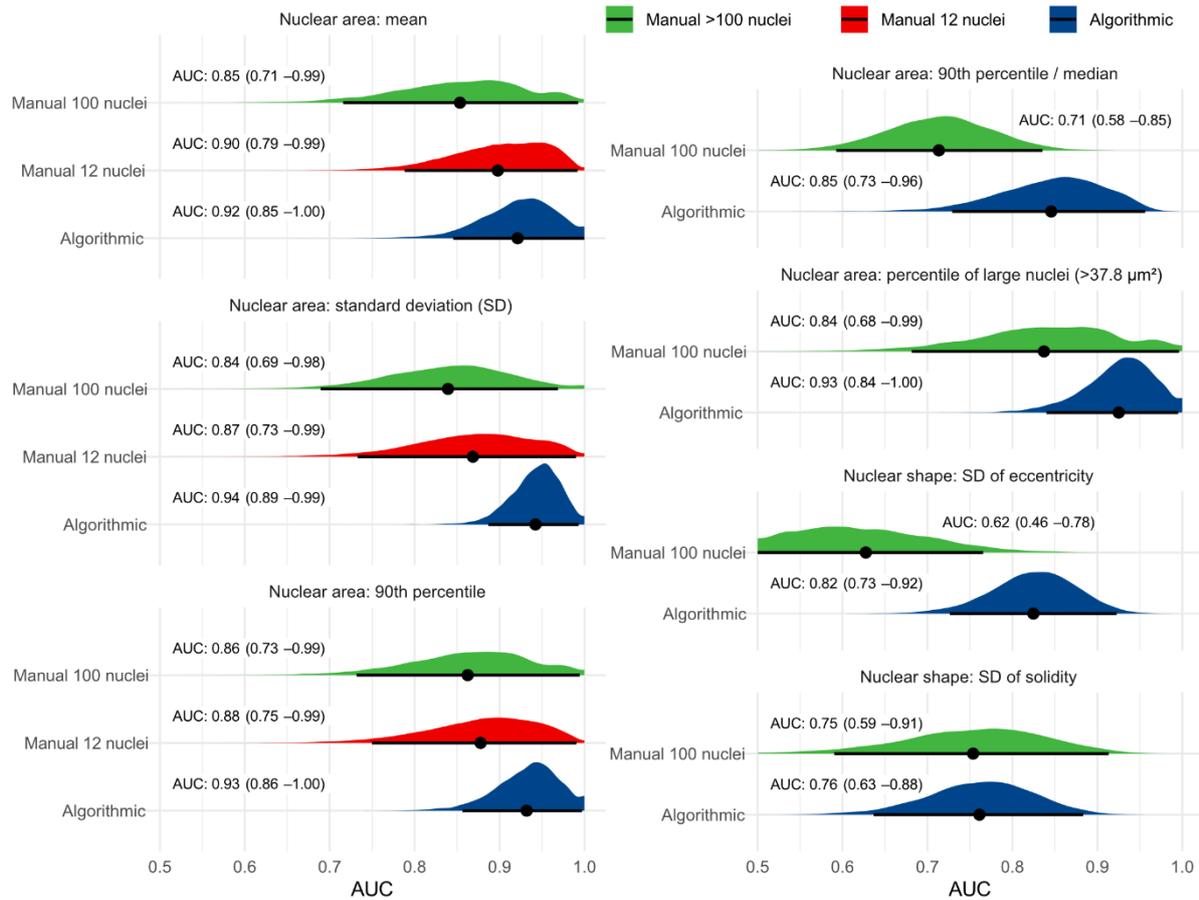

**Figure 4.** Graphical illustration of the area under the curve (AUC; point estimator indicated by black dot) with its 95% confidence intervals (black line) and probability density function (red and blue areas) using bootstrapping for different nuclear morphometry parameters regarding tumor-specific survival. Algorithmic morphometry (Algo) is displayed in blue and manual morphometry by pathologists (Path) in red density functions.

For practicable manual morphometry of 12 nuclei the maximum value is equivalent with the 90th percentile.

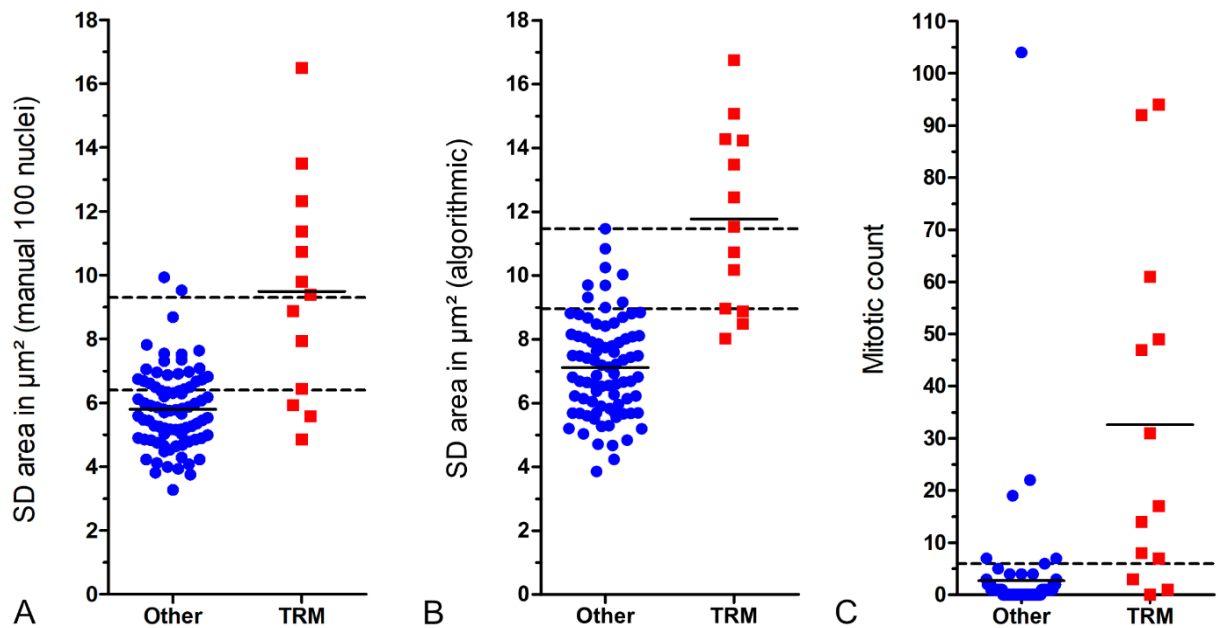

**Figure 5.** Scatterplots comparing cases with tumor-related mortality (TRM) with others (survived follow-up period or died due to tumor-unrelated cause). The short solid lines represent the mean of the measurement of the respective outcome group. A) Standard deviation (SD) of nuclear area measured by gold standard manual morphometry. The lower broken line represents threshold 1 (6.4 µm²; sensitivity: 76.9%, specificity: 71.1%) and the upper broken line represents threshold 2 (9.3 µm²; sensitivity: 53.8%, specificity: 97.6%). B) standard deviation (SD) of nuclear area measured by fully automated morphometry . The lower broken line represents threshold 1 (9.0 µm²; sensitivity: 76.9%, specificity: 89.2%) and the upper broken line represents threshold 2 (11.5 µm²; sensitivity: 53.8%, specificity: 100%).  C) Mitotic count. The broken line represents the threshold according to Romansik et al. [44] (≥6; sensitivity: 76.9%, specificity: 92.8%).

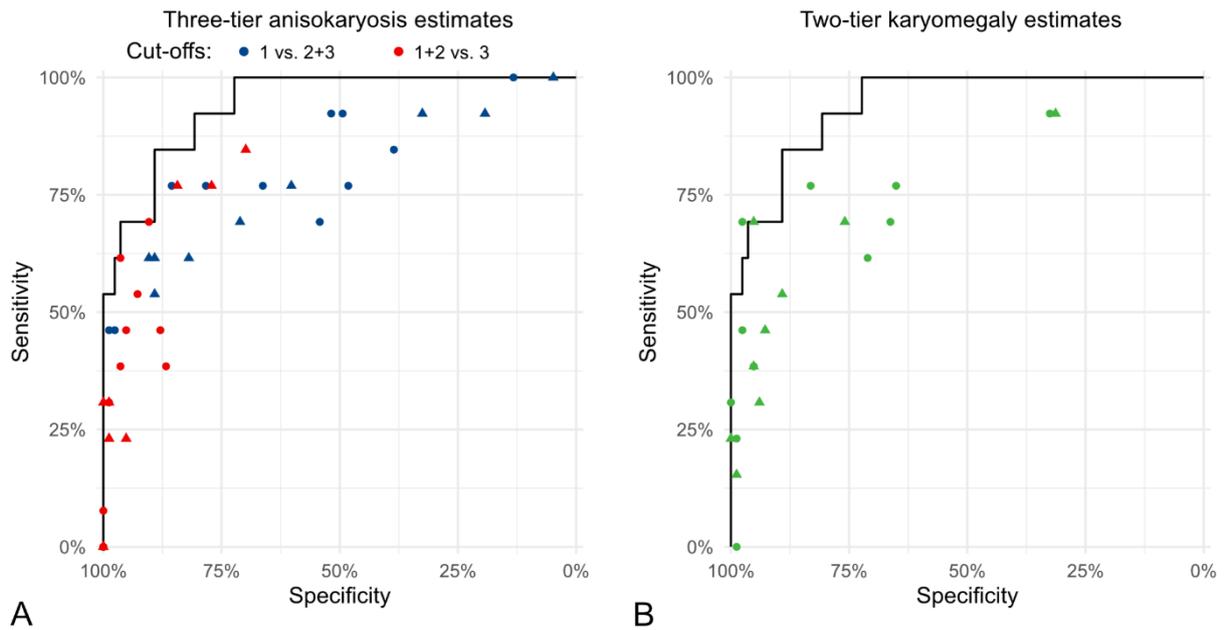

**Figure 6.** Comparison of the pathologists' sensitivity and specificity values for anisokaryosis estimates of time point 1 (dots) and 2 (triangle) regarding tumor-related mortality. The solid line in both graphs represents the same ROC curve for SD area measured by the deep learning-based algorithm. **A)** Pathologists' estimates (dots and triangles) based on the three-tier anisokaryosis (1: none to mild, 2: moderate or 3: severe) approach. Two of the three categories are combined into none to moderate vs. severe (red symbols) and none to mild vs. moderate to severe (blue symbols) anisokaryosis. **B)** Pathologists' estimates by the karyomegaly definition (green symbols).

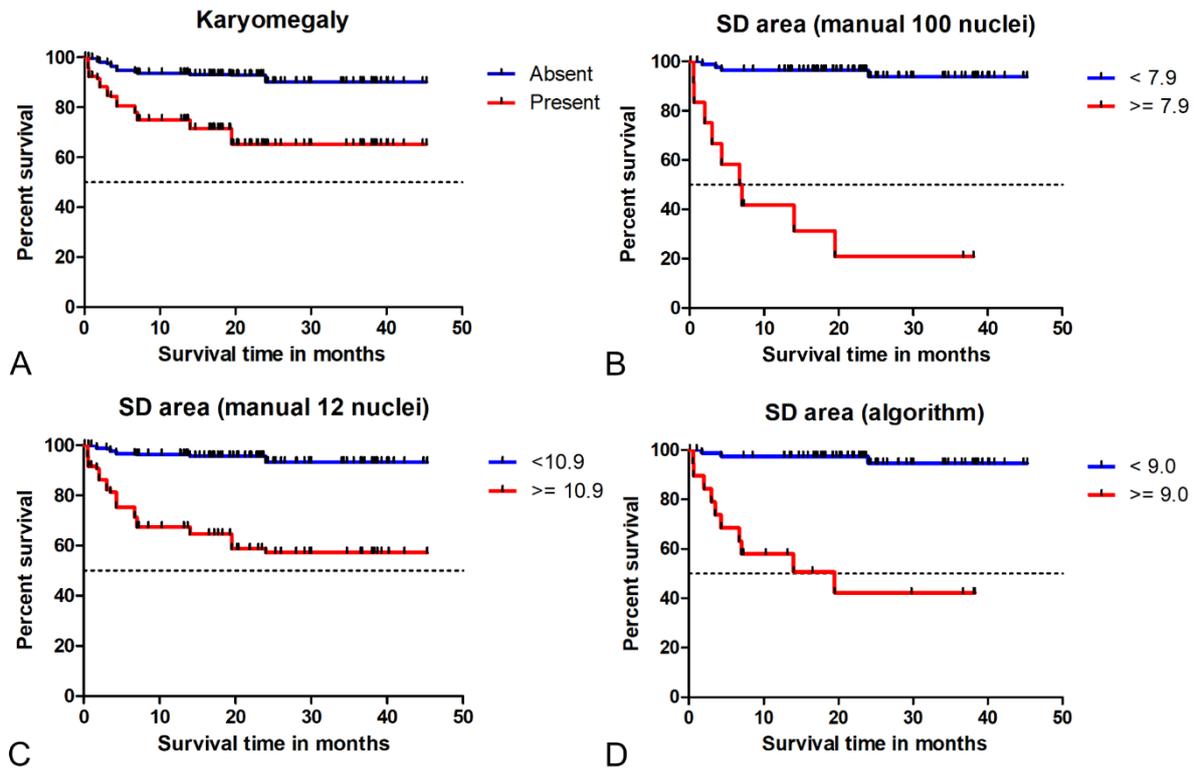

**Figure 7.** Kaplan Meier curves regarding tumor-specific survival time for different tests on nuclear size evaluation. **A)** Karyomegaly estimates (combined data from time point 1 of all 11 pathologists). The hazard ratio is 7.6 (95% CI: 5.7 – 10.1, p < 0.001). **B)** Standard deviation (SD) of nuclear area measured by gold standard manual morphometry (≥100 nuclei). The hazard ratio for this test is 24.8 (95% CI: 7.5 – 81.2, p < 0.001). **C)** SD of nuclear area measured by practicable manual morphometry (12 nuclei; combined data of all 9 pathologists). The hazard ratio for this test is 9.0 (95% CI: 6.0 – 13.4, p < 0.001). **D)** SD of nuclear area measured by algorithmic morphometry. The hazard ratio for this test is 18.3 (95% CI: 5.0 – 67.1, p < 0.001).

**Table 2.** Sensitivity, specificity, and precision regarding tumor-related mortality for the standard deviation (SD) of nuclear area measured by the different morphometric methods. The classification threshold is adapted equally for all tests to the sensitivity value of the mitotic count at the proposed threshold by Romansik et al. of ≥6 (sensitivity: 76.9%; specificity: 92.8%; precision: 62.5%).

| Method | Pathologist | Threshold | Sensitivity | Specificity | Precision |
|---|---|---|---|---|---|
| Manual (≥100 nuclei) | N/A | ≥6.4 µm² | 76.9% | 71.3% | 29.4% |
| Manual (12 nuclei) | P1 | ≥8.0 µm² | 76.9% | 84.3% | 43.5% |
|  | P2 | ≥9.8 µm² | 76.9% | 84.3% | 43.5 % |
|  | P3 | ≥10.9 µm² | 76.9% | 88.0% | 50.0% |
|  | P5 | ≥13.4 µm² | 76.9% | 86.7% | 47.6% |
|  | P6 | ≥8.5 µm² | 76.9% | 69.9% | 28.6% |
|  | P7 | ≥12.2 µm² | 76.9% | 95.2% | 71.4% |
|  | P8 | ≥10.5 µm² | 76.9% | 84.3% | 43.5% |
|  | P10 | ≥14.0 µm² | 76.9% | 90.4% | 55.6% |
|  | P11 | ≥12.2 µm² | 76.9% | 84.3% | 43.5% |
| Algorithmic | N/A | ≥9.0 µm² | 76.9% | 89.2% | 52.6% |

N/A, not applicable; P1 – P11, pathologists 1 - 11

**Intra-tumoral Heterogeneity (Algorithmic Morphometry)**

Some variability in the measurements between the tumor locations of the 3-5 ROIs was noted for the different algorithmic morphometric parameters. Regarding the prognostic classification based on threshold 1 (the sensitive threshold) of SD of the nuclear area, 39/96 cases (41%) had divergent (low and high) measurements between the individual ROIs (Table 3). Based on threshold 2 (the specific threshold), divergent classification between the ROIs of one tumor occurred in 18/96 cases (19%). However, the overall test performance (determined by the AUC) only increased mildly with the number of ROI used for statistical analysis, except for SD of eccentricity.

Tumor heterogeneity as a prognostic test had high AUC value, i.e., cases with a higher number of ROIs above the prognostic threshold or cases with high variability between the different tumor locations were more likely to be aggressive tumors. For the morphometric parameter of SD of nuclear area, the SD between the different ROI had an AUC of 0.790 (95% CI: 0.632 – 0.947), and the proportion of hotspot ROI (above threshold 1 = 9.0 µm²) had an AUC of 0.890 (95% CI: 0.780 – 1.00). With a higher proportion of hotspot ROI, the death probability increases (Table 3), and grouping cases with 0 – 20% vs. ≥40% hotspot ROI resulted in a sensitivity of 92.3% and specificity of 78.3%.

**Table 3.** Distribution of the outcome cases based on the proportion of hotspot regions of interest (ROI, classified by threshold 1) according to the standard deviation of nuclear area measurements (fully automated morphometry) [a] comparing cases with tumor-related mortality (TRM) to other cases without TRM. For each case 3 (N = 1), 4 (N = 4) or 5 (N = 91) ROI were analyzed.

| Outcome | Proportion of hotspot ROI | | | | | | |
|---|---|---|---|---|---|---|---|
| | 0/3-5 (0%) | 1/5 (20%) | 2/5 (40%) | 2/4 (50%) | 3/5 (60%) | 4/5 (80%) | 5/5 (100%) |
| TRM | 1 | 0 | 3 | 0 | 0 | 2 | 7 |
| Other | 47 | 18 | 6 | 1 | 6 | 3 | 2 |
| Death probability | 2% | 0% | 33% | 0% | 0% | 40% | 78% |

[a] calculated by the number of ROI with a measurement above threshold 1 (9.0 µm²) divided by the number of evaluated ROI.

# Discussion

This study assessed several nuclear evaluation methods, including anisokaryosis estimates, manual morphometry, and algorithmic morphometry pertaining to reproducibility, measurement accuracy, and prognostic utility. This work represents the first veterinary study to: 1) compare anisokaryosis estimates with morphometry and 2) develop and validate 2A) a practicable manual morphometry method and 2B) algorithmic morphometry. The results of this study highlight the limitations of estimating nuclear characteristics of tumor cells as part of prognostic histologic evaluation, particularly regarding rater reproducibility. Nuclear morphometry has several advantages over categorical evaluation of nuclear features (see introduction); however, large numbers of cells cannot be measured by pathologists in a routine diagnostic setting. Measurement of ≥100 nuclei (the gold standard) required, an average of 17.5 minutes. Our proposed practicable manual nuclear morphometry and automated nuclear morphometry approaches can overcome these limitations while having a high prognostic value and practicability that allows integration into routine diagnostic workflow of laboratories using digital microscopy. The choice of using small ROIs for algorithmic morphometry was mostly made to ensure practicability of the proposed approach, including the need for computational resources for image analysis. The size of the ROIs is roughly equivalent to microphotographs at 400x using a camera mounted on a light microscope, making this approach also feasible for pathologists that do not have access to whole slide images.

The results of this study show low inter-rater reproducibility for anisokaryosis estimates in ccMCT. Given the wide variation in sensitivity and specificity when comparing individual pathologist's estimates, this parameter has different prognostic value between pathologists. Low inter-rater reproducibility may be the result of the following two issues: 1) distinct anisokaryosis / nuclear pleomorphism categories are difficult to precisely define and 2) the defined thresholds between the categories are interpreted differently by individual pathologists. A vague definition of the categories (i.e. mild, moderate, severe or similar) is common in current literature;[17,18,28,30,39,40,42,46] thus, we decided to use a similar approach for the three-tier anisokaryosis method. Other studies provide a more specific definition for the anisokaryosis categories with percentage of affected cells and/or degree (X-fold) of nuclear size variation,[21,27,35] as for the karyomegaly definition used in this study.

Regardless of the use of a vague or more specific definition, study participants had difficulty applying the categories in the same manner. The specific definitions (X-fold size variation in X% of nuclei) are interpreted differently by pathologists when estimated. Further studies are needed to evaluate methods that improve reproducibility, such as pictorial illustrations of each category or reference sizes within the image based on a scale overlay (digital microscopy) or non-tumor cell within the image (light microscopy), similar to that used for the lymphoma subtype classification.[51] Of note, average neoplastic mast cells in the tumor section seem to be a less ideal reference size, as aggressive tumors not only have a higher variation in nuclear size but also a higher mean/median nuclear area (see results of morphometry). When creating new definitions for reproducible anisokaryosis estimates, the numerical morphometry data may guide finding prognostically meaningful thresholds a priori.

Low reproducibility of pathologists' estimates is the main motivation for morphometry [15,47] and the development of the different algorithmic approaches (see introduction). Manual morphometry has been shown to have high reproducibility,[15] but it is impractical to measure a larger number of cells for routine diagnostic pathology. In our study, we limited the number of cells to 12 (stratified sampling method) with the intention of creating a fairly practical quantitative test, although some study participants commented that they found annotating the 12 nuclei challenging for the time available in a diagnostic setting. Despite improved reproducibility as compared to estimates, significant differences between pathologists were still noted, suggesting that 12 selected nuclei with stratified sampling may not be representative enough for each case. Interestingly, the maximum nuclear area measured by manual morphometry (12 nuclei) had the highest inter-rater reproducibility in this study (compared to anisokaryosis estimates and other parameters of practicable manual morphometry), while also having a good prognostic value (see below). Measuring just one of the largest neoplastic nuclei in the tumor section would improve the feasibility of applying morphometry and should be evaluated as a prognostic factor in future studies. Due to intra-tumoral heterogeneity (see below), reproducibility may be reduced when this task is performed on WSI and not restricted to a few preselected tumor regions. Future studies should compare reproducibility of manual morphometry depending on the number of cells annotated and the method of cell selection (complete vs. stratified). Using one algorithm will result in 100% reproducibility;[11] however, the reproducibility between different segmentation models (based on different network architectures and/or training/validation data) that may be applied in different laboratories needs to be evaluated (inter-algorithmic reproducibility).

Identification of nuclei and their outlines is generally a straightforward task for pathologists and trained deep learning models. However, it is our experience from the manual annotations (pathologists experiments and creation of the ground truth dataset) that identification of the cell type (neoplastic mast cell vs. stromal cell etc.) and nuclear borders can be difficult for some cells. Nuclear membranes may be obscured in heavily granulated ccMCT. We therefore evaluated the measurement accuracy of manual and algorithmic morphometry. Compared to the ground truth, manual morphometry of 12 nuclei resulted in an overestimation of nuclear size, suggesting that pathologists had a tendency to oversample larger cells within the ROI. In contrast, the algorithm sampled most nuclei in the ROIs (little sampling bias) and was able to accurately segment mast cell nuclei, resulting in a smaller error for nuclear size measurements. The perceived sources of errors of the segmentation model area are discussed in the following sentences. Undersegmentation (algorithmic division of the image into too few segments, i.e, nuclei, leading to the interpretation of excessively large nuclei) was a rare problem that resulted in slight overestimation of karyomegalic cells. As compared to other tumor types, undersegmentation may be less relevant due to the lack of close cell contact between the neoplastic round cells and the moderate amount of cytoplasm that separates the neighboring mast cell nuclei. Oversegmentation (segmentation of only a part of the nucleus) and omission of neoplastic cells rarely occurred, particularly when cytoplasmic granules obscured the nucleus. It should be noted that the omission of neoplastic cells does not affect the overall morphometry if this is a random error. Another source of error that rarely occurred for our algorithm is the segmentation of non-neoplastic nuclei, such as stromal cells or endothelial cells, which should be included in higher numbers in future training datasets. It was beyond the scope of the present study to evaluate the robustness of the segmentation models to changes in the image characteristics, particularly the different whole slide image scanners and tumor types, as these have been identified as relevant sources of a domain shift that lead to reduction of the algorithmic performance of deep learning-based mitotic figure algorithms.[3,4]

All these sources of algorithmic error, i.e., undersegmentation, false object localization and potential domain shift, may have a bias on the nuclear size measurements and further improvements of the model are warranted. The potential sources of domain shift (particularly different whole slide image scanners) should be considered for development of future datasets. Deep learning-based models are the state-of-the-art for nuclear segmentation (as compared to traditional machine learning methods) as demonstrated by several computer science challenges.[23,29] Future studies should evaluate an instance segmentation model for ccMCT that can better separate overlapping/connected nuclei and complete segment or omit obscured nuclei. Based on the frequency and extent of error in nuclear segmentation, different degrees of human-machine interaction (computer-assisted vs. fully automated prognosis) may be recommended to reduce errors during case evaluation.[7,38] In this study, we did not apply any human-machine interaction; however, we suggest that the model's segmentation mask (overlay on the HE image) should be verified visually by a trained pathologist if algorithmic morphometry is applied for routine diagnostic service. This transparency of the intermediate results of the segmentation model can be used to remove algorithmic errors, including under- and oversegmentation as well as detection of non-neoplastic nuclei.

Two-dimensional morphometric measurements were employed in the present study. Other studies have used stereological estimates of nuclear volume (three-dimensional).[15,47] The rationale for estimating nuclear volume is that the measured area of the nucleus (a three-dimensional structure) in two-dimensional tissue sections is influenced by the position and orientation of the nucleus to the plane of section.[13,15] The incorrect assumption of orderly positioning and orientation of the nuclei along the plane of section introduces bias, such as the increased chance of evaluating larger nuclei more frequently.[13] While we acknowledge that our measurements may not perfectly correlate with 3D nuclear characteristics, we assume that our measurements follow statistical principles that allow calculation of the probability density function. We argue that the 3D methods might not be ideal either. Calculations of a 3D volume from a 2D area measurement are based on the assumption that all 3D structures have the same volume, while the difference in the measured area is related to the different planes of sections through the nuclei. However, a uniform nuclear volume is not the case for neoplastic nuclei with anisokaryosis. The benefit of 2D morphometry is that different parameters (mean, SD of nuclear area, 90th percentile etc.) of size and shape can be evaluated, while the stereologic approach is restricted to the mean volume. A direct comparison of the prognostic value of two- and three-dimensional-based methods may be an interesting subject for future studies.

Most nuclear size parameters had good prognostic value, while practicable manual and algorithmic morphometry predicted outcome a little bit better than gold standard manual morphometry on average. The prognostic ability of practicable manual morphometry is encouraging considering that this test was restricted to 12 nuclei using a stratified sampling strategy, while the gold standard manual method evaluated 100 nuclei with a complete sampling within random grid fields. Our findings stand in contrast to the conclusions of Casanova et al.,[15] who found limited prognostic value of the volume-weighted mean nuclear volume (three-dimensional morphometry). The benefit of algorithmic morphometry is that a large number of nuclei (usually >1,000, complete sampling strategy) can be evaluated increasing the representativeness of the measurement and thus increasing the prognostic value as compared to the manual measurement.

While most size measurements had a very similar prognostic relevance, the parameters 90th percentile / median (approximating the 2011 karyomegaly definition [27]) was not particularly relevant in this study population due to the high correlation between the 90th percentile and median, which reduced the effect size of the quotient. Defining karyomegaly through the proportion of tumor cells with enlarged nuclei above a specific nuclear area (such as ≥ 38 µm²) seems to be more appropriate; however, it is impractical for evaluations using traditional light microscopy. Of note, the manual measurement of one of the largest nuclei in the image (maximum measurement by pathologists), reflecting the presence or absence of few karyomegalic cells, had surprisingly high prognostic value and may be a very practical prognostic test. Nuclear morphometry may represent a useful alternative to karyomegaly estimates for use in the 2011 grading system,[27] providing that the laboratory uses digital microscopy. Validation of our findings are needed in larger study populations.

Another topic for future research is to combine different nuclear characteristics, such as nuclear size, shape, orientation, and spatial distribution.[1,31] In our study, manual and algorithmic shape assessments had a markedly lower prognostic relevance than nuclear size parameters; however, it was beyond the scope of this study to evaluate whether nuclear shape assessments can add prognostic information when combined with nuclear size parameters. Further studies are needed to evaluate the prognostic relevance of the different morphometric parameters of automated morphometry for further tumor types.

Limitations of the present study are the lack of confirmation of the cause of death of the patients and the low number of cases with tumor-related death. These limitations justify a validation study on a second outcome dataset. The low number per outcome event also hindered multivariate analysis in determining whether a combination of several size and shape parameters had an added value.[54] The prognostic results of the MC are as expected from previous literature,[9,10,12,44] suggesting that this outcome dataset is representative. A great advantage of this outcome population is that all patients were exclusively treated by complete surgical removal eliminating the bias of variable treatment strategies on patient survival.

Tumor heterogeneity is interesting regarding sampling strategies of tumor regions and understanding tumor biology. The use of a deep learning-based algorithm allowed us to analyze several ROI and thus enabled intra-tumoral comparison of morphometric measurements in ccMCT for the first time. Although we observed variability of the morphometric measurements between the different tumor regions, the number of ROI used for prognostic evaluation generally had a minor effect on the determined AUC values. While analysis of a single ROI provided a satisfactory prognostic interpretation of the case, a higher number of ROI or potentially larger ROI sizes might be slightly beneficial, which should be evaluated in a future study in more detail. Of note, the maximum value of the analyzed ROI per case did not result in a higher discriminative ability of patient survival than the average of all ROI. However, as we had selected the representative tumor regions without attention to nuclear features and restricted the analysis to a few ROI, it remains unknown whether areas with the highest morphometric values in the tumor section would be favorable for prognostic assessment of the case.

It is very intriguing that intra-tumoral heterogeneity of morphometry itself had a moderate to good discriminative ability for patient survival, even though the heterogeneity measurement was restricted to 5 tumor regions. It would be interesting to explore the distribution of nuclear parameters throughout the entire tumor sections, as has been previously done for nuclear morphometry in human tumors [1] or for the mitotic count in ccMCT.[7,8] Particularly, the prognostic test "proportion of hotspot ROI" would probably benefit from a more comprehensive analysis of the entire tumor section. These fully automated nuclear morphometry algorithms have great potential for employing intra-tumoral heterogeneity as a prognostic test.

## Conclusion

Poor rater reproducibility of anisokaryosis estimates hinders meaningful application of this test for routine tumor prognostication. An alternative to estimation is nuclear morphometry, the advantages of which include high reproducibility and the capability of determining meaningful prognostic thresholds based on the association with patient outcome. While manual measurements of a larger number of cells is impractical for routine application, we have shown that assessment of a few (12) tumor nuclei using a stratified sampling strategy provides improved rater reproducibility (as compared to estimates) and a meaningful prognostic information (as compared to the gold standard manual morphometry method). Additionally, we propose a deep learning-based algorithm that is capable of analyzing thousands of cells within seconds at low computational costs. This study demonstrated a high measurement accuracy and high prognostic value of fully automated nuclear morphometry for ccMCT. In this study population, morphometric parameters evaluating the nuclear area (such as the standard deviation or $90^{th}$ percentile) were particularly prognostically relevant. The results of this study encourage application of automated nuclear morphometry for routine tumor evaluation in laboratories with established digital workflows. A more thorough investigation of tumor heterogeneity and its prognostic value is warranted based on our preliminary findings with a relatively low number of tumor regions.

## Acknowledgement

Parts of this research was conducted for the diploma project of Eda Parlak using a preliminary version of the DL-based algorithm.

## Declaration of conflict of interest

## Funding

The author(s) received no financial support for the research, authorship, and/or publication of this article.

## Authors' Contributions

CAB, AH, EP, TAD, and AB designed the experiments; MK, CAB, TAD and RK contributed histological sections; MK provided outcome information; EP developed the ground truth dataset; AH developed the algorithm; HJ, MA, SMW, and JS contributed to algorithm development and evaluation; CAB and JG organized the pathologist experiment; TAD, CAA, PB, MJD, AFB, AK, KJ, RK, SM, BR, and FYS participated as study pathologists (anisokaryosis estimates and practicable manual nuclear morphometry); CAB performed mitotic counts and gold standard manual morphometry; CAB, AH, AB, and EP performed data analysis; the manuscript was written by CAB and AH with contributions by all authors.